\documentclass[10pt,twocolumn,letterpaper]{article}

\usepackage{subfigure}
\usepackage{graphicx}
\usepackage{cvpr}
\usepackage{times}
\usepackage{epsfig}
\usepackage{graphicx}
\usepackage{amsmath}
\usepackage{amssymb}
\usepackage{enumerate}
\usepackage{multirow}
\usepackage{cite}

\usepackage[breaklinks=true,bookmarks=false]{hyperref}

\cvprfinalcopy 

\def\cvprPaperID{****} 

\begin{document}

\title{Cross-Age LFW: A Database for Studying Cross-Age Face Recognition in Unconstrained Environments}

\author{Tianyue Zheng, Weihong Deng and Jiani Hu\\
Beijing University of Posts and Telecommunications\\
Beijing 100876,China\\
{\tt\small 2231135739@qq.com, whdeng@bupt.edu.cn, 40902063@qq.com}
}

\maketitle

\begin{abstract}
   Labeled Faces in the Wild (LFW) database has been widely utilized as the benchmark of unconstrained face verification and due to big data driven machine learning methods, the performance on the database approaches nearly 100\%. However, we argue that this accuracy may be too optimistic because of some limiting factors. Besides different poses, illuminations, occlusions and expressions, cross-age face is another challenge in face recognition. Different ages of the same person result in large intra-class variations and aging process is unavoidable in real world face verification. However, LFW does not pay much attention on it. Thereby we construct a Cross-Age LFW (CALFW) which deliberately searches and selects 3,000 positive face pairs with age gaps to add aging process intra-class variance. Negative pairs with same gender and race are also selected to reduce the influence of attribute difference between positive/negative pairs and achieve face verification instead of attributes classification. We evaluate several metric learning and deep learning methods on the new database. Compared to the accuracy on LFW, the accuracy drops about 10\%-17\% on CALFW.
\end{abstract}

\section{Introduction}
Face recognition is a popular topic in computer vision for its wide range of applications and there are two widely used paradigms of face recognition: identification and verification. Face verification attempts to verify whether the given two face images represent the same person or two different people. It is often assumed that neither of the photos shows a person from any previous training set.

Labeled Faces in the Wild (LFW) database \cite{LFWTech} has been widely used as benchmark to study face verification. To form the dataset, they used photos collected as part of the Berkeley Faces in the Wild project \cite{1315253,Berg2004Whos}. Photos in the project were collected from the Yahoo News during 2002 to 2003 and were captured in uncontrolled environments with a wide variety of settings, poses, expressions and lightning. These pictures were popular for researches but because of more than 10\% noisy labels and large number of duplicates, these pictures could not be used as benchmark. So in \cite{LFWTech} they manually cleaned the data, designed new protocols and released the dataset named 'Labeled Faces in the Wild'. The LFW database includes 13,233 face images of 5,749 individuals and there are two views of LFW for experiments including view 1 for development purpose and view 2 as a benchmark for comparison. In view 2, the dataset were separated into 10 non-repeating subsets of images pairs for cross validation. Each subset contains 300 positive pairs (images from the same person) and 300 negative pairs (images from different people). When the database is used only for testing, all the pairs are included (3000 positive pairs and 3000 negative pairs) to obtain the performance results. Since then, hundreds of papers have been published to pursue better performance upon this benchmark in some respect.

 In the recent literature, deep learning shows extraordinary effectiveness in face verification problem due to its superior ability in learning a series of nonlinear feature mapping functions directly from raw pixels. Taigman et al.\cite{Taigman_2014_CVPR} achieved an accuracy of 97.35\% on LFW using large outside training data and deep neural network. Then, Sun et al. \cite{NIPS2014_5416, DBLP:journals/corr/SunLWT15} conducted a series of Deepid and reached an accuracy of 99.53\%. The results have already beyond the performance of human which is 99.20\% on the funneled faces. And Schroff et al. \cite{Schroff_2015_CVPR} developed Facenet and reached 99.63\% on LFW which reported only 22 errors on the entire 6000 images pairs.

 While many deep learning methods have reached nearly saturated accuracy on the benchmark standard Labeled Faces in the Wild (LFW), researchers only solve part of the face verification problem in real world situation. Through inspecting LFW database, one can find two limiting factors which can be improved to better simulate real world face verification. One is that the two images of a positive pair are of similar ages. Although the positive pairs in LFW have different poses and expressions, the images are sometimes taken in the same occlusion and the age gap between positive pairs are often small while the age gap between negative pairs are often large. Due to the different age gap distribution between positive pairs and negative pairs, age estimation accuracy can be a large influence on the performance of LFW face verification. Also, the negative pairs in LFW are selected randomly, which results in different attributes including gender and race of the two people. However, the images of positive pairs are from the same person and have same gender and race. The same/different attributes make the face verification problem more like an attribute classification problem.

\begin{figure}
\centering
\includegraphics[width=0.7\linewidth]
                {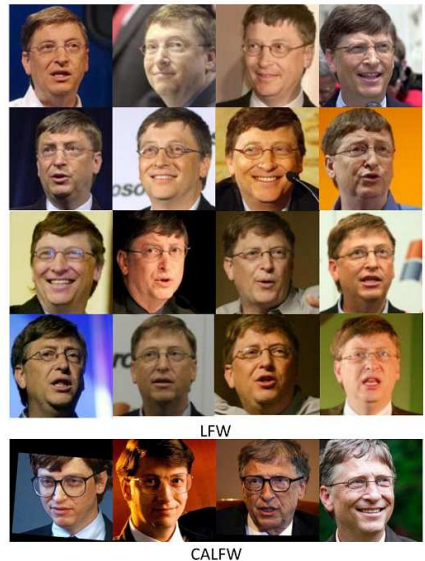}
\caption{The comparison of the same individual in LFW and CALFW. Though LFW has more pictures, aging process in CALFW is more obvious.}
\label{fig:compare_images}
\end{figure}

 In this paper, we consider breaking the two limitation factors of LFW. We reinvent the LFW database by searching images of identities in LFW with apparent age gaps to form positive pairs and selecting negative pairs using individuals with the same race and gender. The new database, called Cross-Age LFW (CALFW) is collected by crowdsourcing efforts to seek the pictures of people in LFW with age gap as large as possible on the Internet so we can add age intra-class variation to the original LFW. After searching, age estimation algorithm \cite{Rothe_2015_ICCV_Workshops} is applied to estimate the age of all the selected pictures and the pairs with largest age gaps are chose as positive pairs in View 2. The database can be viewed and downloaded at the following web address: \url{http://www.whdeng.cn/CALFW/CALFW.html}.
 The comparison of the same person in LFW and CALFW is shown in Figure \ref{fig:compare_images} and from the picture we can see that aging process in CALFW is more obvious.

  We name the new dataset Cross-Age LFW, the prefix "Cross-Age" suggests that age gap of the same individual has been considered as a crucial intra-class variation which better simulates real world face verification situation. Though the images are different from those in LFW, we use the same identities of each fold in LFW and maintain the verification protocols which means our database is a extension of LFW, so the name of our database still includes LFW. There are three motivations behind the construction of CALFW benchmark as follows:

 \begin{itemize}
   \item Establishing a relatively more difficult database to evaluate the performance of real world face verification so the effectiveness of several face verification methods can be fully justified.
   \item Continuing the intensive research on LFW with more realistic consideration on aging process intra-class variation and fostering the research on cross-age face verification in unconstrained situation. The challenge of LFW benchmark mainly focuses on different poses, lightning and expressions in positive pairs while CALFW emphasizes age gap to further enlarge intra-class variance. Also, negative pairs are deliberately selected to avoid different gender or race. CALFW consider both the large intra-class variance and the tiny inter-class variance simultaneously.
   \item Maintaining the data size, the face verification protocol which provides a 'same/different' benchmark and the same identities in LFW, so one can easily apply CALFW to evaluate the performance of face verification.
 \end{itemize}

\section{Related Works}
Face recognition is a popular problem in computer vision for many reasons. First, it is easy to formulate well-posed problem and collect the data we need since individuals come with their name labels. Second, it is worth studying because it is a protruding example of fine-grained classification, the intra-class variation caused by different poses, expressions and ages can often exceed the inter-class variation between two different people. Finally, face recognition problem is of great importance and can be applied to wide ranges of scenarios. For all these reasons, face recognition has become an area which is popular in the vision community.

Typically, there are two types of tasks for face recognition. One is face identification which means that given gallery set and query set, for a given image in the query set, we want to find the most similar face in gallery set and use the identity of the similar face as the identity of the query image. The other type of face recognition is face verification determining whether two given images belong to the same person.

Early face datasets were almost collected under controlled environments such as PIE \cite{1004130}, FERET \cite{879790} and a very high performance can be obtained on these constrained datasets. However, most models learned from these datasets do not work well in practical applications due to the complexity of faces in real world situation. To improve the generalization of face recognition methods, the interests of datasets gradually changed from controlled environment to uncontrolled environment. And so a milestone dataset Labeled Faces in the Wild (LFW) \cite{LFWTech} was established in 2007. Compared to the benchmark dataset before, the biggest difference of LFW is that the images were obtained from Internet rather than acquired under several pre-defined environments. Due to the uncontrolled environment, LFW has various poses, illuminations, expressions, resolutions and these factors are gathered in random way.

Recently, several new face recognition database has been collected to study face recognition and verification. These included CASIA database \cite{DBLP:journals/corr/YiLLL14a}, Megaface \cite{DBLP:journals/corr/MillerKS15}, IJB-A \cite{Klare_2015_CVPR} and FaceScrub \cite{7025068}. The CASIA dataset \cite{DBLP:journals/corr/YiLLL14a} consists of 494,414 images of 10,575 subjects and it is always used to train the deep network. The FaceScrub dataset \cite{7025068} contains 107,818 images of 530 celebrities collected from the web. Each person has an average number of nearly 200 images. Though the percentage of correct labels is difficult to know, these large and deep databases are useful for researchers to train complex face recognition system with complex framework.

 Except for the databases used for training, new protocols and benchmarks have also been proposed for face recognition problem. MegaFace \cite{DBLP:journals/corr/MillerKS15} was designed to study large scale face recognition. The goal of this dataset is to evaluate the performance of current face recognition algorithms with up to a million distractors. Images were derived from the Yahoo 100 Million Flicker creative commons data set \cite{Shamma2016YFCC100M}. All of the images in Megaface were first registered in a gallery with one image each person. Then for each subject in FaceScrub \cite{7025068}, one image was used in the gallery and the rest of the images of the person were used as testing images in an identification paradigm. So the goal of face recognition task was to identify the only one matching images in the 1,000,001 individuals. The dataset was established because many applications require accurate identification at plenty scale. It emphasises the ability to identify individuals in very large galleries, or in the open set recognition problem.
 IJB-A dataset \cite{Klare_2015_CVPR} was introduced to push the frontiers of unconstrained face detection and recognition. The database contains 500 individuals with manually localized face images. It is a mix of images and videos which contains full pose variation and can be joint used for face recognition and face detection. The dataset supports both face recognition and face verification.

 Many datasets have been designed to measure the performance using criteria that are more strict than that of LFW. For verification, the verification rates at 0.1\% false acceptance rate. For identification, rank-1 recognition accuracy on a gallery of millions of people is designed. These protocols and datasets may also involve many comparisons between different ages of the same person. However, the age gap occurs due to large amount data of the same person, rather than human operation. In addition, while simulating more realistic performance, these new databases lose the feature of LFW as the easy-to-use, low barriers to entry. In contrast, we manually add age variations to the same person to enlarge intra-class variations while at the same time using the people with same race and gender as negative pairs to avoid attribute difference influence of positive pairs and negative pairs in face verification. Meanwhile, we design the database by strictly following the protocols of LFW so that researchers need not to do any changes when using the new dataset. These characteristics make the proposed CALFW database totally different from those datasets above.

\section{From LFW to Cross-Age LFW}
Our benchmark is used to achieve face verification. To simulate real world face recognition situation, we add age difference of the same person into the dataset while keeping the identities of LFW at the same time. In this section, we describe the process of the construction of CALFW from collecting data to forming training and testing set in detail.
\subsection{Consruction Details}
 The process of building CALFW dataset can be broken into the following steps:
 \begin{enumerate}
   \item Gathering raw images from the Internet
   \item Running a face detector and manually correcting the results when there are more than one person in the picture
   \item Cropping and rescaling the detected faces
   \item Eliminating duplicate picture
   \item Judging whether labels are correct
   \item Obtaining landmarks and aligning images
   \item Estimating the age of each image and forming pairs of training and testing sets. Selecting the largest age gap pairs as positive pairs and the people with same gender and race as negative pairs.
 \end{enumerate}

 \textbf{Gathering Images.}

 In order to collect images from a large number of people, Google, Bing and GettyImages are utilized to search face images using the identities in LFW dataset. In order to collect images across different ages, we argue the celebrity name with adjectives such as "young", "old", "childhood" as key words.
 The identities in LFW are divided into 300 groups randomly with 40 identities in each group (there are repeated individuals between different groups). Each subject is searched by at least two students to guarantee that the subject has been searched carefully. 300 volunteers who are Chinese students of 18-22 years old have taken part in the collecting mission and they are asked to find two images of each person with age gap as large as possible. Considering the fact that finding more than one picture for some individuals are difficult, the volunteers need to find at least 25 subjects pairs among the 40 individuals.

 \textbf{Detecting Faces.}

 The next step is detecting face, a recently published deep learning tools \cite{7553523} is applied to detect faces in the pictures. For each detection result, the following procedure is performed:
 \begin{enumerate}
   \item If the region is determined by the detector to be a non-face, it is omitted from the database.
   \item If the detecting result shows that there are more than one face in the picture, we manually choose which we need.
   \item If the detecting result detects the face we need, the face is cropped and rescaled (as described below) and saved as a separate JPEG file.
 \end{enumerate}

\textbf{Cropping and rescaling.}

 For those images placed in CALFW dataset, we use the following procedure to create them. The region obtained by face detector for the given face is expanded by 2 according to the maximum value of length and width. If the expanded region falls outside the original region of the image, a new image of the size equal to the size we want will be created by using black pixels to fill in the area outside the original image. The expanded image is then resized to $250 \times 250$ using the Matlab function imresize. Finally the image is saved in JPEG format.

 \textbf{Eliminating duplicate face photos.}

 Before removing the duplicate images, we need to define what is duplicates. The simplest definition is that the two images are numerically equivalent at each pixels. However, the definition ignores many situations when faces in the images are indistinguishable to the human eyes for the reason that the images collected by volunteers might have been recropped, rescaled, renormalized or variably compressed. Thus if we do not eliminate these face photos, we might form positive pairs which are visually equivalent but differed numerically. So according to \cite{LFWTech}, we choose to define duplicates as images which are judged to have a common original source photograph. To remove duplicate images, we have the following two steps. First, a structural similarity measure \cite{1284395} is used to compare all the possible couples of images from the same identity. Only the couples with a very high similarity are inspected and we delete the low quality version. To make sure that there is no duplicate image in the dataset, we then manually check pictures of each individual.

 \textbf{Judging whether labels are correct.}

 For each given subject, we pay extreme cautious to manually judge the scraped images to be truly about this celebrity or not. We use the images in the standard LFW as reference and whenever we are not sure about the label, we will try to find the original web page of the scraped image and read the page content to guide the label. The rich information of the original page benefits the quality of labeling, especially for those hard cases. When the identity of the image can not be confirmed by web page, three judging people see the image together and get the final decision based on voting result.  In total, we have more than 10,000 images label which spent many hours. While we attempt to label all the pictures correctly, it is possible that certain people have been given incorrect names.

 \textbf{Obtaining landmarks and aligning images.}

 To achieve face alignment, we need to obtain face landmarks first. We first use \cite{7553523} to obtain basic landmarks and then adjust them manually to guarantee accurate landmarks positions (including two eyes and the midpoint of mouth). Then, we use the coordinates of the three landmarks to align the CALFW original images. The three landmarks are best fit to a set of predefined "average" coordinates via a similarity transform. Considering that similarity transform only accounts for rotation and scale, faces of different poses and shapes differ in the aligned coordinates of the fiducial points but the face looks natural and does not have distortion.

 \textbf{Forming training and testing sets.}

 In LFW view 2, it defines 10 disjoint subsets of image pairs which are suitable for cross validation. Each subset contains 300 positive pairs and 300 negative pairs. The 10 subsets are organized by their identities and each identity only appears once in certain subset. Based on it, our CALFW dataset has been divided into 10 separate folds using the same identities contained in the LFW 10 folds. The CALFW dataset contains 4,025 individuals with 2,3 or 4 images for each person. To estimate the age of each image, we use Dex \cite{Rothe_2015_ICCV_Workshops} which is the winner of the ChaLearn LAP 2015 challenge on age estimation and the name of each image is formed as follows:
 $$name\_0001.jpg,  name\_0002.jpg,$$
 $$name\_0003.jpg,  name\_0004.jpg,$$
 the number "0001" "0002" reflects the rank of age estimation result. "0001" is the youngest image and "0004" is the oldest image of a subject.

 We then form training and testing sets. There are two rules for constructing positive pairs. First, we want the pairs to contain people as many as possible to reflect the variety in real world face verification, so if the number of people in the fold is larger than 300, one positive pair is selected for each individual. Second, to reflect the age gap feature in our dataset, we select positive pairs with the largest age gap for each individual. When it comes to negative pairs, to avoid the attribute difference of positive pairs (same gender and race) and negative pairs (random race and gender), we first manually label the race and gender of each person in CALFW and then select negative pairs with people who have the same gender and race randomly.

\subsection{Comparison between LFW and CALFW}
In this section, we compare the standard LFW and our CALFW. For the reason that the characteristic of our database is the age variation for same person, we first compare the age gap between LFW and CALFW, we then compare the attribute (gender, race and age gap) between positive pairs and negative pairs in LFW and CALFW and finally compare the number of images for each person.
\begin{figure*}
\centering
\includegraphics[width=1.0\linewidth]
                {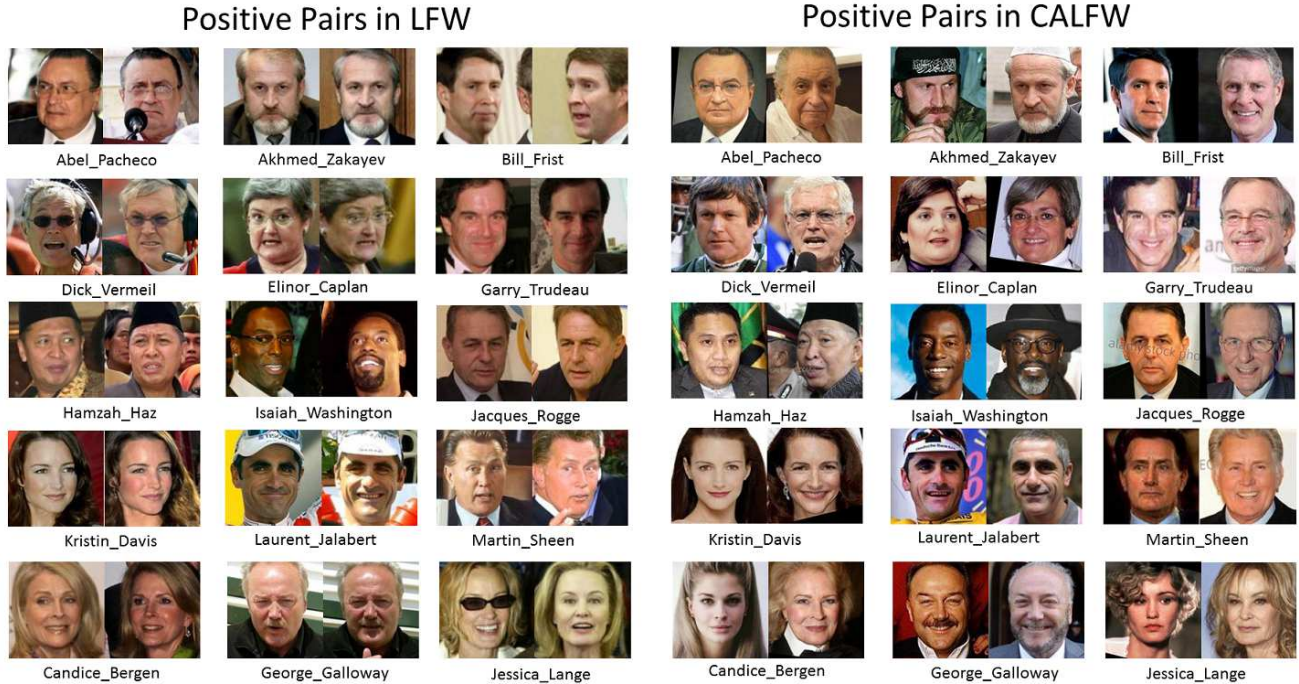}
\caption{The comparison of positive pairs in LFW and CALFW. Compared to LFW, the positive pairs in CALFW contain obvious age difference.  }
\label{fig:img_positive}
\end{figure*}

\begin{figure*}
\centering
\includegraphics[width=1.0\linewidth]
                {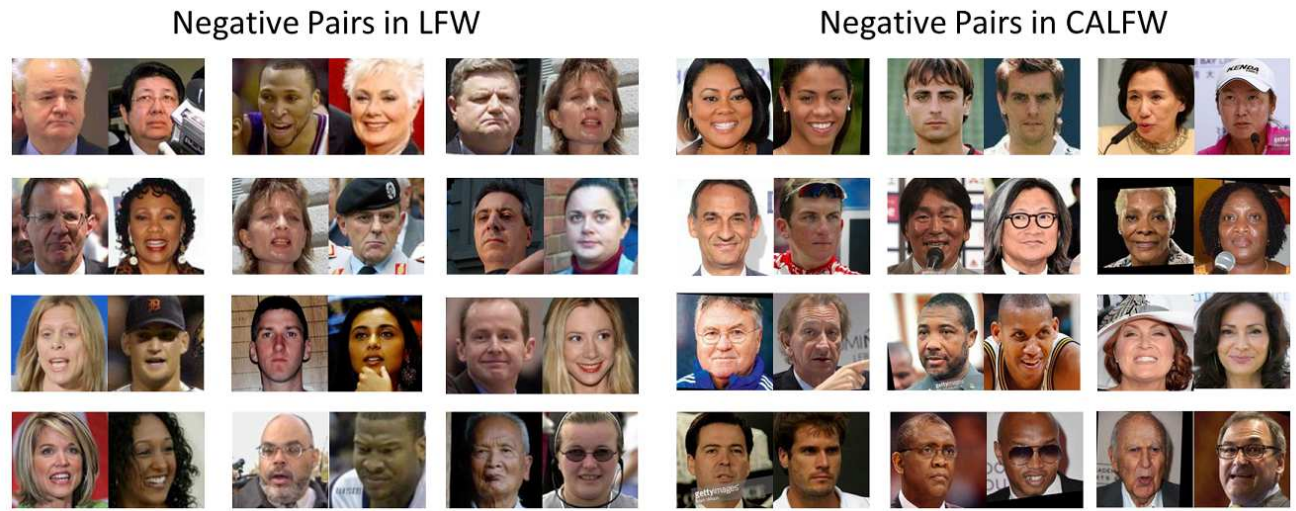}
\caption{The comparison of negative pairs in LFW and CALFW. Compared to LFW, the negative pairs in CALFW have same gender and race, which reduces the influence of attribute difference between positive pairs and negative pairs in face verification. }
\label{fig:img_negative}
\end{figure*}
\begin{figure}
  \centering
  \subfigure[The age gap distribution of positive pairs and negative pairs in LFW.]{
    \label{fig:subfig:a} 
    \includegraphics[width=0.4\textwidth]{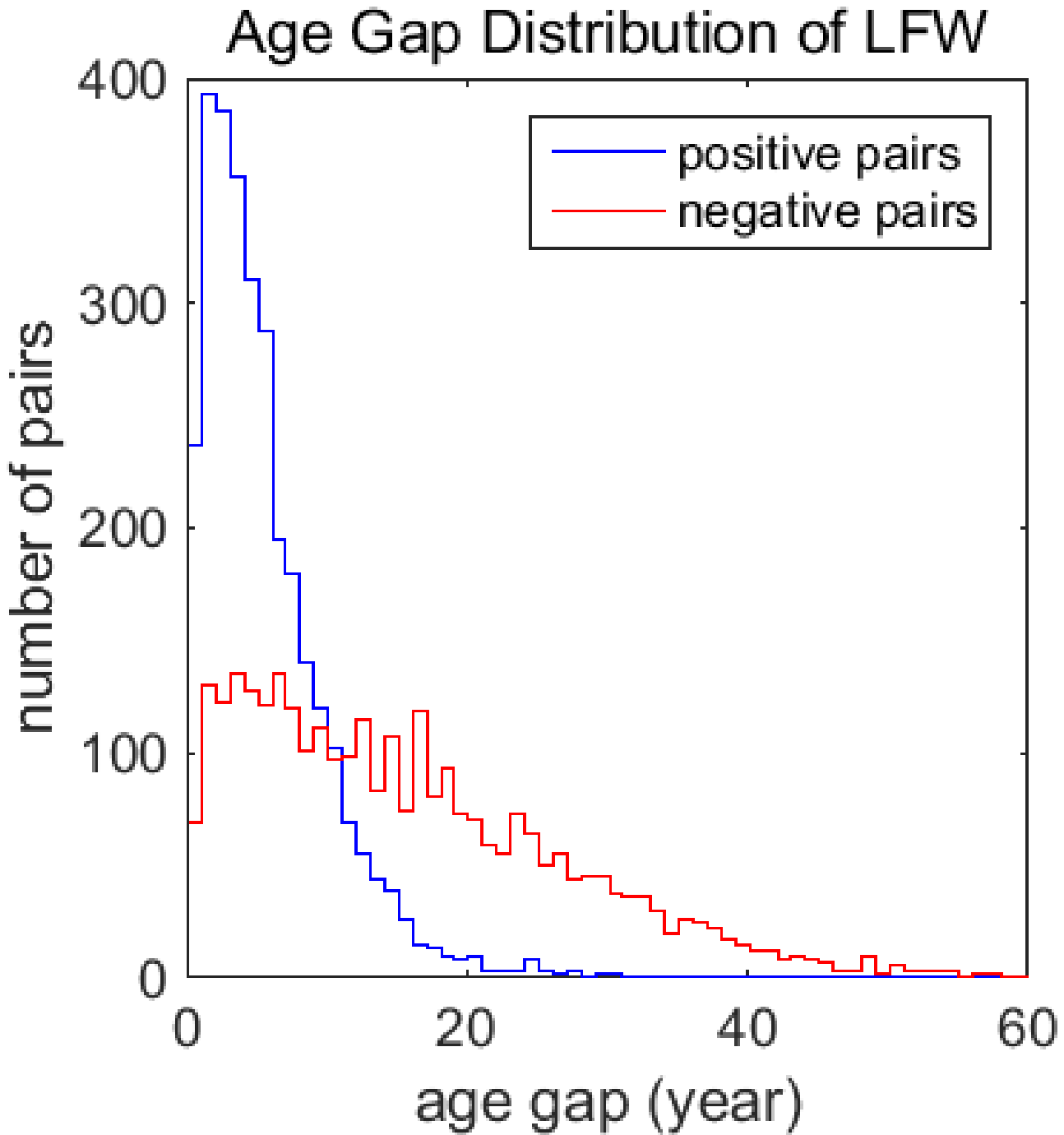}}
  \hspace{1in}
  \subfigure[The age gap distribution of positive pairs and negative pairs in CALFW.]{
    \label{fig:subfig:b} 
    \includegraphics[width=0.4\textwidth]{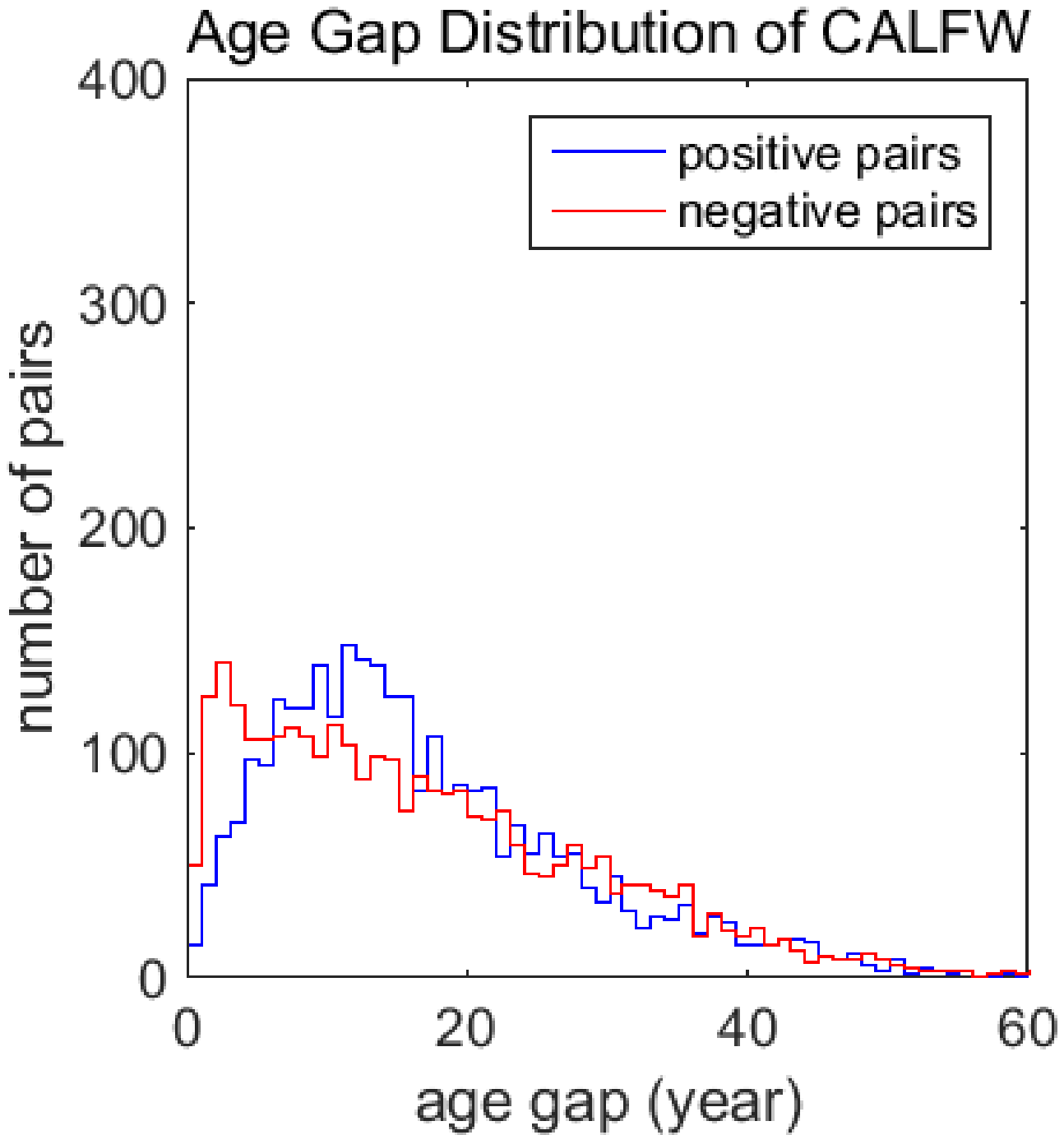}}
  \caption{ Compared to the positive pairs in LFW, the age gaps of positive pairs in CALFW is larger. This shows we successfully add aging process to intra-class variations. Also, in LFW, age gaps of most positive pairs are less than 10 years while that of most negative pairs are larger than 10 years, in CALFW, there is no clear boundary to distinguish the two kinds of pairs, so age gap can not be a big influence on face verification in CALFW.}
  \label{fig:age_lfw_calfw} 
\end{figure}

To compare age gap, we first use Dex \cite{Rothe_2015_ICCV_Workshops} to estimate the age gap of pairs in View 2. The average age gap and standard deviation of positive pairs and negative pairs in LFW and CALFW are shown in Table \ref{tab:ave_age} and the age gap distributions are shown in Figure \ref{fig:age_lfw_calfw}. For LFW, the average age gap of positive pairs is 4.94 while that of negative pairs is 14.85. According to the distribution, we have the following observations. First, standard LFW to some extent ignores the aging process of people though age has a large impact on intra-class variations. Second,age gaps of the two kinds of pairs are different. Age gap of most positive pairs are less than 10 years while most negative pairs are larger than 10 years, which means that age gap classification (e.g. using 10 years as boundary) can influence the results of face verification to some extent.
\begin{table}
\caption{Age gap average value of positive pairs and negative pairs in LFW and CALFW.}
\begin{center}
\begin{tabular}{|c|c|c|}
\hline
 Pairs& LFW &CALFW \\
\hline\hline
 Positive Pairs & 4.94$\pm$4.24 & 16.61$\pm$10.78\\
 Negative Pairs & 14.85$\pm$11.00 & 16.14$\pm$11.88 \\
\hline
\end{tabular}
\end{center}
\label{tab:ave_age}
\end{table}

The age gap mean value and distribution of CALFW are reported in Table \ref{tab:ave_age} and Figure \ref{fig:age_lfw_calfw}. Compared to the distribution of LFW positive pairs, the age gap distribution of CALFW positive pairs crosses larger age gap range and CALFW has larger positive pairs age gap mean value, which confirms that CALFW considers aging process to enlarge the intra-class variance. Also, the age gap distribution of positive and negative pairs is more similar to each other and we can not find a boundary to classify whether the pair is positive or not. This shows that age gap attribute can not be a feature to distinguish positive pairs and negative pairs in CALFW.

Then we compare the attribute (gender, race and age gap) between positive pairs and negative pairs in LFW and CALFW. Considering all the negative pairs are randomly selected in LFW, it is common that two individuals have different gender or race while the positive pairs are of the same race and gender. The age gap difference, gender difference and race difference make face verification problem to some extent like an attribute classification problem. To confirm this, we design an attributes feature for each pair in LFW and CALFW. We use $g \epsilon \{0,1\}$ to represent whether the two individuals in each pair are of the same gender, $r \epsilon \{0,1\}$ to represents whether the pair are of the same race and $a$ to represents the age gap of each pair estimated before. The attributes feature of each pair can be expressed as $\{g,r,a\}$.
Three criteria are applied to classify negative pairs and positive pairs. The first is that if two images are of different gender, the pair is negative pair. The second is that if two images are of different race, the pair is negative pair. The third is that if the age gap is larger than 10 years, the pair is negative pair. If not, the pair is positive pair. We use the attributes feature and the strategy to classify positive/negative pairs in LFW and the result reaches an accuracy of 86.23\% according to Table \ref{tab:attri_class}. The result shows that the big attribute difference of positive pairs and negative pairs in LFW influences face verification problem. If a method is effective for face attribute classification task, it can obtains a relatively good result in LFW face verification task.

To avoid attribute influence in face verification problem, we select negative pairs according to race and gender attributes in CALFW. We first label the gender and race of each person and then randomly form negative pairs using people with the same gender and race.
We still use attributes feature and the same classification criteria to achieve face verification task but this time to adapt the different age gap distribution and obtain better classification result, we adjust the age boundary to 26 and the results are shown in Table \ref{tab:attri_class}.
The accuracy of face verification in CALFW reaches 51.80\% which is almost random. Compared to 86.23\% in LFW, the accuracy drops sharply. The result indicates that there is little attribute (including age gap, gender and race) difference between positive pairs and negative pairs of CALFW thus the performance of face verification in CALFW can not be effected by attribute difference and it can fully reflects the ability of methods proposed to handle face verification problem.
\begin{table}
\caption{Face verification accuracy using attributes feature.}
\begin{center}
\begin{tabular}{|c|c|}
\hline
 Dataset& Accuracy \\
\hline\hline
 LFW & 86.23\% \\
 CALFW &51.80\%\\
\hline
\end{tabular}
\end{center}
\label{tab:attri_class}
\end{table}
\begin{figure}
\centering
\subfigure[Image number of each person in LFW.]{
\label{fig:subfig:c} 
\includegraphics[width=2.5in]{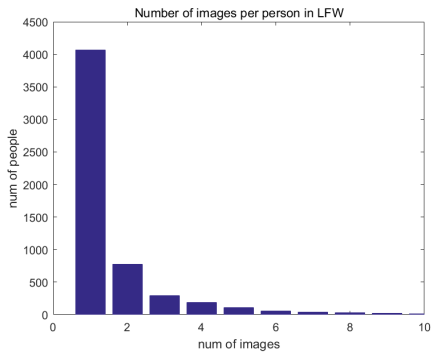}}
\hspace{1in}
\subfigure[Image number of each person in CALFW.]{
\label{fig:subfig:d} 
\includegraphics[width=2.5in]{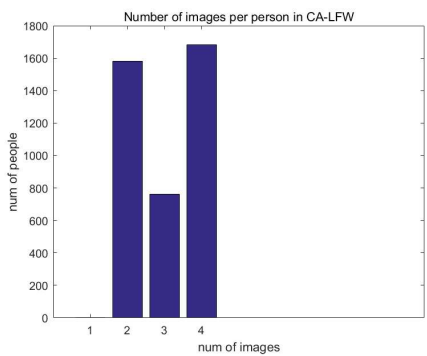}}
\caption{The image number of each person in LFW and CALFW. Compared to LFW, the number distribution of CALFW is more balanced and the positive pairs contain more people.}
\label{fig:num_lfw_calfw} 
\end{figure}

The comparison of positive pairs and negative pairs in LFW and CALFW are shown in Figure \ref{fig:img_positive} and Figure \ref{fig:img_negative}, the pictures show that compared to LFW, the positive pairs in CALFW contain obvious age difference and the negative pairs in CALFW have same gender and race, which reduces the influence of attribute difference between positive pairs and negative pairs in face verification.

Also, we notice that in LFW, the image number of each person is not balanced. The database contains images of 5,749 individuals while 4,096 people have just a single image which means they can only be used in negative pairs. To increase the number of people in positive pairs so that the limited 3000 positive pairs can better reflect the diversity of face verification in real world face recognition, each individual in CALFW has at least 2 images. The distributions of each person's image number in LFW and CALFW are shown in Figure \ref{fig:num_lfw_calfw}. For the purpose of display, we only show people with 10 images or less considering the fact that only a small number of people have more than 10 pictures in LFW.

In conclusion, there are three main differences between LFW and CALFW. First is that age difference has been added to intra-class variations in CALFW to better simulate real-world face verification. Second is that instead of randomly selecting negative pairs, to avoid the influence of attributes difference of positive and negative pairs, we select negative pairs using people with the same gender and race. Third is that the image number of each person in CALFW and LFW. CALFW is more balanced with 2,3 or 4 images for each person while the distribution of CALFW is not balanced. As a result of it, CALFW contains more people in positive pairs.

\section{Baseline}
In this section, we briefly introduce some well-established methods for face verification problem. Considering that LFW has largely promoted the research on metric learning and deep learning, we compare the performance of these methods between LFW and the proposed CALFW benchmark. This comparison helps us understand how difficult the CALFW dataset is.

\subsection{Comparison on metric learning.}

We first evaluates the metric learning methods designed for unconstrained face verification and successfully tested on LFW. The common thinking of metric learning approaches is to learn a good distance function to reduce the distance of positive pairs and enlarge the distance of negative pairs simultaneously. Specially, we test Euclidean distance, Mahalanobis distance, Information Theoretic Metric Learning (ITML) \cite{Davis2007Information}, Keep It Simple and Straightforward Metric Learning (KISSME) \cite{Roth2012Large}, Support Vector Machine (SVM) \cite{Cortes1995Support} under image-restricted protocol, in which only 6000 face pairs are available.
We apply Local Binary Patterns (LBP) \cite{1717463} as feature for the comparison on metric learning approaches. Specially, we use the images in LFW and CALFW aligned by similarity transform and extract 59-bin uniform pattern LBP histogram in each of the ($8 \times 15$) non-overlapping blocks of the facial images. To reduce the dimension of LBP, they are projected into a 300-dimensional PCA subspace before metric learning.
\begin{table}
\caption{Comparison of mean verification accuracy(\%) on LFW and CALFW under image-restricted setting using LBP feature.}
\begin{center}
\begin{tabular}{|c|c|c|}
\hline
 Approach& LFW &CALFW \\
\hline\hline
 Euclidean distance & 74.68$\pm$0.92 & 62.37$\pm$1.79\\
 Mahalanobis distance & 75.05$\pm$1.94 &66.83$\pm$1.09\\
 SVM \cite{Cortes1995Support}& 76.68$\pm$1.19 & 65.27$\pm$1.96\\
 ITML \cite{Davis2007Information}&  82.37$\pm$2.05 &68.82$\pm$1.36\\
 KISSME \cite{Roth2012Large}& 77.57$\pm$1.85 & 67.87$\pm$0.93\\
\hline
\end{tabular}
\end{center}
\label{tab:acc_lbp}
\end{table}
\begin{figure}
\centering
\includegraphics[width=1.0\linewidth]
                {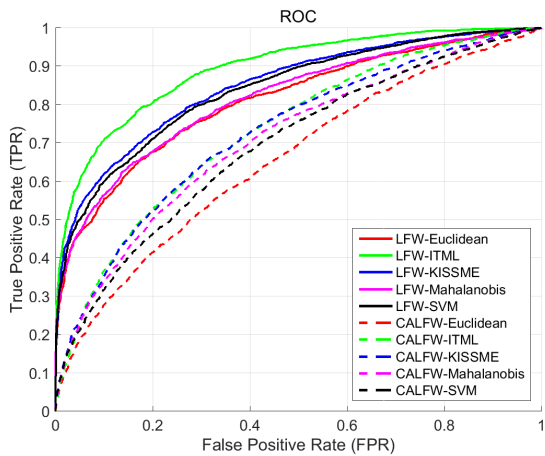}
\caption{The ROC curves of various metric learning methods on LFW and CALFW using LBP feature.}
\label{fig:roc_lbp}
\end{figure}
Mean verification accuracy of LBP feature using metric learning methods under the image-restricted protocols are compared in Table \ref{tab:acc_lbp} with the ROC curves shown in Figure \ref{fig:roc_lbp}. ITML method reaching an accuracy of 82.37\%, which is the beat mean accuracy of the 10 cross-validation criteria. However, the accuracy drops to 68.82\% on CALFW and the performance of other approaches also drop sharply. KISSME drops about 10\% and SVM drops about 11\%.

\subsection{Comparison on deep learning approaches}
\begin{figure}
\centering
\includegraphics[width=1.0\linewidth]
                {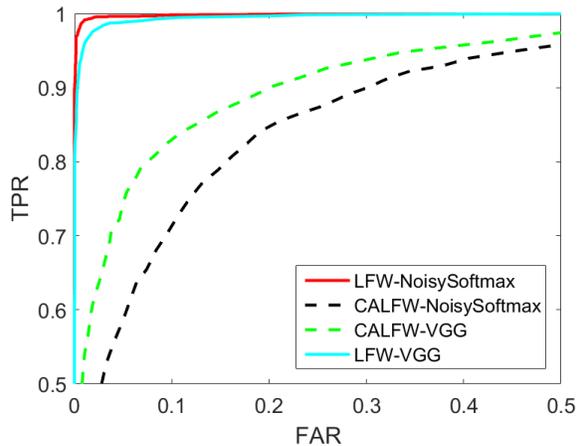}
\caption{The ROC curves of various deep learning methods on LFW and CALFW.}
\label{fig:dplearn_roc}
\end{figure}

Recently, deep convolutional neural networks trained by massive labeled outside data have reported fairly good results on face verification task of LFW benchmark. Due to the good performance, we apply two well-published convolutional neural networks to compare LFW and CALFW, they are VGG-Face \cite{Parkhi2015Deep} and Noisy Softmax \cite{Chen_2017_CVPR}. The parameters of two networks are set according to the original papers and we apply the network model directly. The VGG-Face descriptors are extracted using the off-the-shelf CNN model based on the VGG-Very-Deep-16 CNN architecture as described in \cite{Parkhi2015Deep}. The comparison of face verification accuracy on LFW and CALFW are reported in Table \ref{tab:acc_dplearn} and the corresponding ROC curves are shown in Figure \ref{fig:dplearn_roc}.

\begin{table}
\caption{Comparison of mean verification accuracy(\%) on LFW and CALFW using deep learning approaches.}
\begin{center}
\begin{tabular}{|c|c|c|}
\hline
 Approach& LFW &CALFW \\
\hline\hline
 VGG-Face \cite{Parkhi2015Deep}& 97.85\% & 86.50\%\\
 Noisy Softmax \cite{Chen_2017_CVPR}& 99.18\% &82.52\%\\
\hline
\end{tabular}
\end{center}
\label{tab:acc_dplearn}
\end{table}

According to the accuracy results and the ROC curves, deep learning approaches perform better than metric learning methods using LBP feature, which reflects the effectiveness of convolutional neural networks methods to describe images. VGG achieves 97.85\% on LFW and the accuracy drops about 11\% on CALFW. For Noisy Softmax network, the accuracy drops form 99.18\% to 82.52\%, which shows that by adding age gap to intra-class variations and using negative pairs with the same gender and race, the dataset becomes difficult for face verification.

\section{Discussion and Future work}
In this paper, we have constructed a novel dataset according to the well-established LFW to develop face verification: the Cross-Age Labeled Faces in-the-Wild (CALFW) collection. The main contributions of the proposed database are: First, collecting new images according to the identity list of LFW so that each individual contains at least 2 images in the dataset. Due to the balanced distribution, more people are involved to form positive pairs to simulate the diversity of intra-class variations in real world face verification. Second, our benchmark focuses on age gap intra-class variation rather than common face discrimination. Third, we concern at the attribute difference of positive pairs and negative pairs. The images of positive pairs in LFW often have similar ages, same gender and same race, while the randomly selected negative pairs are often with large age gaps, different gender and race. To narrow the attributes difference, we randomly select people with same gender and race as negative pairs.
Finally, the benchmark described in this paper provides a unified testing protocol which can easily evaluate human face verification.

We test the validity of our database by using a convolution neural network age estimation tool and report baseline performance achieved by metric learning and deep learning methods. Empirical results suggest that the CALFW dataset provides new challenge for face verification.

Our benchmark focuses on age gap of the same person while the similarly-lookings of different people are not investigated. To increase the difficulty of face verification, the age gap of positive pairs and the looking similarity of negative pairs can be combined together to form comprehensive face verification problem. We look forward to exciting research inspired by our dataset and benchmark.


{\small
\bibliographystyle{ieee}
\bibliography{egbib}

\begin{thebibliography}{10}\itemsep=-1pt

\bibitem{1717463}
T.~Ahonen, A.~Hadid, and M.~Pietikainen.
\newblock Face description with local binary patterns: Application to face
  recognition.
\newblock {\em IEEE Transactions on Pattern Analysis and Machine Intelligence},
  28(12):2037--2041, Dec 2006.

\bibitem{Berg2004Whos}
T.~L. Berg, A.~C. Berg, J.~Edwards, and D.~A. Forsyth.
\newblock Whos in the picture.
\newblock pages 264--271, 2004.

\bibitem{1315253}
T.~L. Berg, A.~C. Berg, J.~Edwards, M.~Maire, R.~White, Y.-W. Teh,
  E.~Learned-Miller, and D.~A. Forsyth.
\newblock Names and faces in the news.
\newblock In {\em Proceedings of the 2004 IEEE Computer Society Conference on
  Computer Vision and Pattern Recognition, 2004. CVPR 2004.}, volume~2, pages
  II--848--II--854 Vol.2, June 2004.

\bibitem{Chen_2017_CVPR}
B.~Chen, W.~Deng, and J.~Du.
\newblock Noisy softmax: Improving the generalization ability of dcnn via
  postponing the early softmax saturation.
\newblock In {\em The IEEE Conference on Computer Vision and Pattern
  Recognition (CVPR)}, July 2017.

\bibitem{Cortes1995Support}
C.~Cortes and V.~Vapnik.
\newblock Support-vector networks.
\newblock {\em Machine Learning}, 20(3):273--297, 1995.

\bibitem{Davis2007Information}
J.~V. Davis, B.~Kulis, P.~Jain, S.~Sra, and I.~S. Dhillon.
\newblock Information-theoretic metric learning.
\newblock In {\em International Conference on Machine Learning}, pages
  209--216, 2007.

\bibitem{LFWTech}
G.~B. Huang, M.~Ramesh, T.~Berg, and E.~Learned-Miller.
\newblock Labeled faces in the wild: A database for studying face recognition
  in unconstrained environments.
\newblock Technical Report 07-49, University of Massachusetts, Amherst, October
  2007.

\bibitem{Klare_2015_CVPR}
B.~F. Klare, B.~Klein, E.~Taborsky, A.~Blanton, J.~Cheney, K.~Allen,
  P.~Grother, A.~Mah, and A.~K. Jain.
\newblock Pushing the frontiers of unconstrained face detection and
  recognition: Iarpa janus benchmark a.
\newblock In {\em The IEEE Conference on Computer Vision and Pattern
  Recognition (CVPR)}, June 2015.

\bibitem{DBLP:journals/corr/MillerKS15}
D.~Miller, I.~Kemelmacher{-}Shlizerman, and S.~M. Seitz.
\newblock Megaface: {A} million faces for recognition at scale.
\newblock {\em CoRR}, abs/1505.02108, 2015.

\bibitem{7025068}
H.~W. Ng and S.~Winkler.
\newblock A data-driven approach to cleaning large face datasets.
\newblock In {\em 2014 IEEE International Conference on Image Processing
  (ICIP)}, pages 343--347, Oct 2014.

\bibitem{Parkhi2015Deep}
O.~M. Parkhi, A.~Vedaldi, and A.~Zisserman.
\newblock Deep face recognition.
\newblock In {\em British Machine Vision Conference}, pages 41.1--41.12, 2015.

\bibitem{879790}
P.~J. Phillips, H.~Moon, S.~A. Rizvi, and P.~J. Rauss.
\newblock The feret evaluation methodology for face-recognition algorithms.
\newblock {\em IEEE Transactions on Pattern Analysis and Machine Intelligence},
  22(10):1090--1104, Oct 2000.

\bibitem{Roth2012Large}
P.~M. Roth, P.~Wohlhart, M.~Hirzer, M.~Kostinger, and H.~Bischof.
\newblock Large scale metric learning from equivalence constraints.
\newblock In {\em Computer Vision and Pattern Recognition}, pages 2288--2295,
  2012.

\bibitem{Rothe_2015_ICCV_Workshops}
R.~Rothe, R.~Timofte, and L.~Van~Gool.
\newblock Dex: Deep expectation of apparent age from a single image.
\newblock In {\em The IEEE International Conference on Computer Vision (ICCV)
  Workshops}, December 2015.

\bibitem{Schroff_2015_CVPR}
F.~Schroff, D.~Kalenichenko, and J.~Philbin.
\newblock Facenet: A unified embedding for face recognition and clustering.
\newblock In {\em The IEEE Conference on Computer Vision and Pattern
  Recognition (CVPR)}, June 2015.

\bibitem{Shamma2016YFCC100M}
D.~A. Shamma, D.~A. Shamma, G.~Friedland, B.~Elizalde, K.~Ni, D.~Poland,
  D.~Borth, and L.~J. Li.
\newblock Yfcc100m: the new data in multimedia research.
\newblock {\em Communications of the Acm}, 59(2):64--73, 2016.

\bibitem{1004130}
T.~Sim, S.~Baker, and M.~Bsat.
\newblock The cmu pose, illumination, and expression (pie) database.
\newblock In {\em Proceedings of Fifth IEEE International Conference on
  Automatic Face Gesture Recognition}, pages 46--51, May 2002.

\bibitem{NIPS2014_5416}
Y.~Sun, Y.~Chen, X.~Wang, and X.~Tang.
\newblock Deep learning face representation by joint
  identification-verification.
\newblock In Z.~Ghahramani, M.~Welling, C.~Cortes, N.~D. Lawrence, and K.~Q.
  Weinberger, editors, {\em Advances in Neural Information Processing Systems
  27}, pages 1988--1996. Curran Associates, Inc., 2014.

\bibitem{DBLP:journals/corr/SunLWT15}
Y.~Sun, D.~Liang, X.~Wang, and X.~Tang.
\newblock Deepid3: Face recognition with very deep neural networks.
\newblock {\em CoRR}, abs/1502.00873, 2015.

\bibitem{Taigman_2014_CVPR}
Y.~Taigman, M.~Yang, M.~Ranzato, and L.~Wolf.
\newblock Deepface: Closing the gap to human-level performance in face
  verification.
\newblock In {\em The IEEE Conference on Computer Vision and Pattern
  Recognition (CVPR)}, June 2014.

\bibitem{1284395}
Z.~Wang, A.~C. Bovik, H.~R. Sheikh, and E.~P. Simoncelli.
\newblock Image quality assessment: from error visibility to structural
  similarity.
\newblock {\em IEEE Transactions on Image Processing}, 13(4):600--612, April
  2004.

\bibitem{DBLP:journals/corr/YiLLL14a}
D.~Yi, Z.~Lei, S.~Liao, and S.~Z. Li.
\newblock Learning face representation from scratch.
\newblock {\em CoRR}, abs/1411.7923, 2014.

\bibitem{7553523}
K.~Zhang, Z.~Zhang, Z.~Li, and Y.~Qiao.
\newblock Joint face detection and alignment using multitask cascaded
  convolutional networks.
\newblock {\em IEEE Signal Processing Letters}, 23(10):1499--1503, Oct 2016.

\end{thebibliography}
}

\end{document}